\documentclass{IOS-Book-Article}

\usepackage{mathptmx}
\usepackage{soul}\setuldepth{article}
\def\hb{\hbox to 11.5 cm{}}

\usepackage{graphicx}
\usepackage{xcolor}
\usepackage{enumitem}
\usepackage{soul}
\usepackage{xspace}
\usepackage{url}
\usepackage{csquotes}
\usepackage{booktabs}
\usepackage{tcolorbox}
\usepackage{subcaption}

\usepackage{hyperref} 
\hypersetup{
    colorlinks=true,    
    linkcolor=blue,     
    citecolor=blue,     
    urlcolor=blue       
}

\definecolor{forthcomingcolour}{RGB}{189,0,38}
\newcommand{\forthcoming}[1]{\textcolor{forthcomingcolour}{{#1}}}

\begin{document}

\footnotetext[0]{%
\forthcoming{Author Accepted Manuscript. Published version: De Sabbata, S., Mizzaro, S. and Roitero, K. Geospatial mechanistic interpretability of large language models. In Janowicz, K. et al. editors. Geography according to ChatGPT. Frontiers in artificial intelligence and applications. IOS Press; \emph{forthcoming}.}
}

\pagestyle{headings}
\def\thepage{}
\begin{frontmatter}              

\title{Geospatial Mechanistic Interpretability of Large Language Models}

\markboth{}{May 2025\hb}

\author[A]{\fnms{Stef} \snm{De Sabbata}\orcid{0000-0002-2750-7579}%
\thanks{Corresponding Author: Stef De Sabbata.}},
\author[B]{\fnms{Stefano} \snm{Mizzaro}\orcid{0000-0002-2852-168X}}, and
\author[B]{\fnms{Kevin} \snm{Roitero}\orcid{0000-0002-9191-3280}}

\runningauthor{De Sabbata et al.}
\address[A]{University of Leicester, UK}
\address[B]{University of Udine, Italy}

\begin{abstract}
Large Language Models (LLMs) have demonstrated unprecedented capabilities across various natural language processing tasks. Their ability to process and generate viable text and code has made them ubiquitous in many fields, while their deployment as knowledge bases and ``reasoning'' tools remains an area of ongoing research. In geography, a growing body of literature has been focusing on evaluating LLMs' geographical knowledge and their ability to perform spatial reasoning. However, very little is still known about the internal functioning of these models, especially about how they process geographical information.

In this chapter, we establish a novel framework for the study of geospatial mechanistic interpretability -- using spatial analysis to reverse engineer how LLMs handle geographical information. Our aim is to advance our understanding of the internal representations that these complex models generate while processing geographical information -- what one might call ``how LLMs think about geographic information'' if such phrasing was not an undue anthropomorphism.

We first outline the use of probing in revealing internal structures within LLMs. We then introduce the field of mechanistic interpretability, discussing the superposition hypothesis and the role of sparse autoencoders in disentangling polysemantic internal representations of LLMs into more interpretable, monosemantic features. 
In our experiments, we use spatial autocorrelation to show how features obtained for placenames display spatial patterns related to their geographic location and can thus be interpreted geospatially, providing insights into how these models process geographical information. We conclude by discussing how our framework can help shape the study and use of foundation models in geography.
\end{abstract}

\end{frontmatter}
\markboth{May 2025\hb}{May 2025\hb}

\section{Introduction}

In the past few years, the field of Artificial Intelligence (AI) has seen an unprecedented pace of change, from the introduction of the transformers architecture in 2017 to the release of ChatGPT in 2022. The ability of Large Language  Models (LLMs) to produce human-like (at least superficially) answers to questions took many by surprise. However, two years forward, the technology had already become an off-the-shelf commodity, and new research streams have emerged that focus on exploring the capabilities of these models in a broad range of specific tasks and subjects. 
In geography, several authors have explored the ability of LLMs to
process geographical information \cite{mai2023opportunitieschallengesfoundationmodels, doi:10.1080/17538947.2024.2353122}, including their ability to perform a broad range of spatial tasks \cite{Hochmair_2024, xu2024evaluatinglargelanguagemodels} and infer correct answers to spatial reasoning questions \cite{cohn2023dialecticallanguagemodelevaluation, https://doi.org/10.4230/lipics.cosit.2024.28, li2024advancing}, resolve toponyms \cite{Hu24092024},
assist practitioners in GIS \cite{li2023autonomous, zhang2023geogptunderstandingprocessinggeospatial, zhang2024bb}, cartography \cite{doi:10.1080/15230406.2024.2404868}, earth observation \cite{singh2024geollm} and urban planning \cite{tan2023promiseschallengesmultimodalfoundation, zhu2024plangptenhancingurbanplanning}, 
process social media posts during natural disasters \cite{doi:10.1080/13658816.2023.2266495} and visual geographical information such as street view and satellite images \cite{Roberts_2024_CVPR}, 
understand urban spaces \cite{feng2024citygptempoweringurbanspatial}, discern intercardinal directions \cite{fulman2024distortionsjudgedspatialrelations}, provide recommendations for points of interest \cite{feng2024move} or routing \cite{ilyankou2024sentencetransformerslearnquasigeospatial}, while other authors have focused on the quality \cite{ roberts2023gpt4geolanguagemodelsees, 10.1145/3589132.3625625}, bias \cite{decoupes2024evaluation}, and diversity \cite{agile-giss-5-38-2024, liuMakingGeographicSpace}
of LLMs' geographical knowledge. However, our understanding of the internal functioning of LLMs is very limited, especially when it comes to geographical information.

Due to the semantic footprints of places in textual content \cite{BERRAGAN2024102121} and LLMs being trained as contextual next-word predictors, it seems likely that an LLM's reaction to placenames referring to places which are nearby or otherwise related to each other might be somehow similar, at least in some facets of the complex internal representation that these model construct when processing information. For instance, both Wikipedia pages about Loughborough\footnote{\url{https://en.wikipedia.org/wiki/Loughborough}} and Market Harborough\footnote{\url{https://en.wikipedia.org/wiki/Market_Harborough}} mention ``market town'' in their first paragraph and ``Leicester'' in their second paragraph. As LLMs have been trained on content from the web, including Wikipedia, it stands to reason that receiving the words ``Loughborough'' or ``Market Harborough'' as input would crank millions of an LLM's internal ``levers'' in position to heighten the probabilities of one of the next words being ``market town'' or ``Leicester'' -- unless ``Loughborough'' is followed by ``Lake''\footnote{\url{https://en.wikipedia.org/wiki/Loughborough_Lake}}, but we will get to placename ambiguity further below. 

\begin{figure}[tb]
    \centering
    \begin{tabular}{c}
     \includegraphics[width=.99\linewidth]{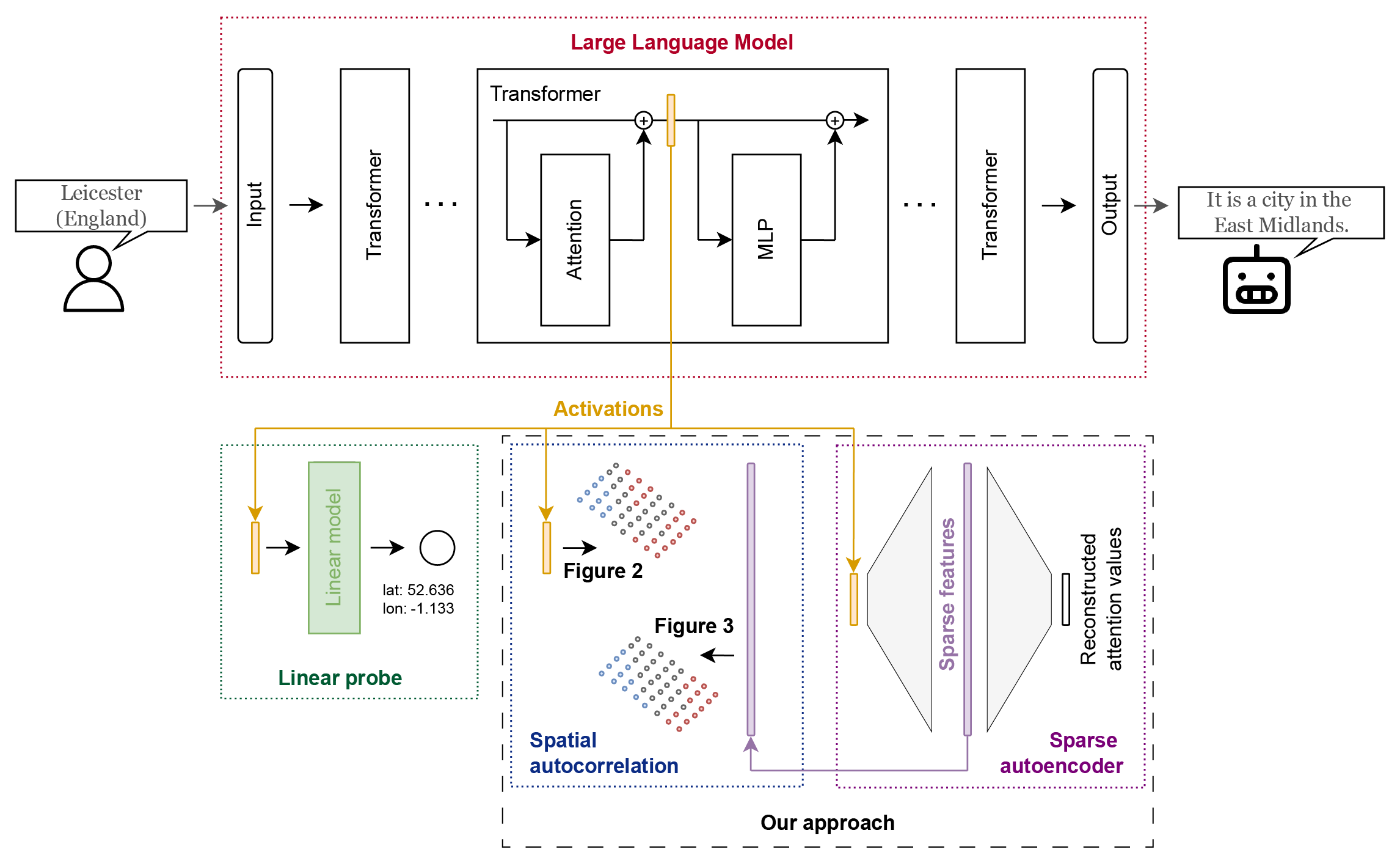}
    \end{tabular}
    \caption{
    An example illustrating the extraction of the activations from an LLM (top); the use of the activations in a linear probe to predict the latitude and longitude of the place mentioned in the input (bottom-left) and a sparse autoencoder (bottom-right); and the use of spatial autocorrelation to analyse the activations and the sparse features (centre). Our approach encompasses the latter two components.
    }
    \label{fig:framework}
\end{figure}

To clarify the terminology used in the rest of the chapter -- a neural network can be thought of as composed of \emph{neurons} laid out in subsequent \emph{layers} with \emph{edges} connecting neurons at one layer to neurons at the next layer \cite{lecun2015deep}. Each edge contains one of the ``levers'' mentioned above, which are called \emph{parameters} (or \emph{weights}) and control the flow of information through the network. Each neuron of the first layer corresponds to a piece of the input (e.g., a pixel of an image, a column of a table, or a word in a sentence). Each piece of information of the input is pushed forward from its neuron at the first layer through the edges towards the second layer. In transit, each edge uses its own parameter to modify the input into a new value, which arrives at the neuron at the second layer. That neuron combines together all the values from all its incoming edges and applies an \emph{activation function} (usually a non-linear function such as a sigmoid or a ReLU \cite{nair2010rectified}) to produce a value commonly referred to as \emph{activation}. The \emph{transformer layers} \cite{vaswani2023attentionneed}, of which LLMs are composed, are sophisticated blocks composed of several layers, including a mechanism called \emph{attention}, which allows LLMs to better understand how different parts of a text influence each other's meaning. The analyses in this chapter focus on the activations extracted after the attention mechanism, before the \emph{Multi-Layer Perceptron (MLP)} component of the transformer layer, as illustrated in Figure~\ref{fig:framework}. A set of activations from the same layer is called an internal \emph{representation} \cite{6472238} of the input at that specific layer, and if it is identified as something humanly interpretable -- such as high output values when the input  references to the Golden Gate Bridge in a language model \cite{templeton2024scaling} -- it is called a \emph{feature} of the input.

The activations are then passed onward to the subsequent layer until the last output layer is reached. During training, the output is compared to the expected output using a \emph{loss function}, and \emph{backpropagation} \cite{lillicrap2020backpropagation} is used to change the parameters to (hopefully) achieve lower loss (i.e., error) at the next training step. The set of steps necessary to train a model once on all the data is called an \emph{epoch}. The number of epochs necessary to train a model depends on factors such as the size and diversity of the training dataset, the size and architecture of the model, the type and difficulty of the task, and the intended qualities of the model, such as accuracy and generalisability. LLMs are initially trained to predict the next word in a sentence, in the phase commonly referred to as pre-training, before the model is fine-tuned for a specific task, such as being a chatbot. Finally, many LLMs go through a phase of reinforcement learning with human feedback, where they are trained to produce the desired type of answers based on positive or negative feedback from human judgements \cite{ziegler2020finetuninglanguagemodelshuman}. Once the training process is completed, the LLM parameters (the ``levers'') are fixed, and an input sentence will pass through the neural network, generating internal representations at each layer until the last layer selects the next word to add to the output.

This chapter focuses on the analysis of those internal representations and the question of whether training an LLM (at its core) as a next-word prediction might lead them to follow patterns akin to  Tobler's ``first law of geography'' \cite{1aa639e4-d759-3f3c-b072-bba8376952da}, where \emph{input text about near things might produce internal representations that are more similar than input text about distant things}. More concretely, we can start from the assumption that a high activation value at a specific neuron indicates that the model has activated specific paths through the neural network to increase the probability of certain output tokens. As the model has been trained on large amounts of human-generated text, including references to and descriptions of places, we can assume that the model has encoded information about places, their geographies and characteristics. Therefore, we can formulate the following conjecture: close or otherwise similar places are likely to be associated with similar LLM internal representation patterns. 

Recent studies \cite{lietard2021language, gurneeLanguageModelsRepresent2024, godeyScalingLawsGeographical2024, chen2023correlationlargelanguagemodels} were able to use linear probes to find broad, continental and national patterns of latitude and longitude (as discussed in Section \ref{sec:probing}). However, linear probes are limited in their ability to capture geographical relationships. Therefore, to better understand LLMs' internal representation of geographical information, we focused our study on a null hypothesis rooted in spatial analysis: if an LLM has encoded information about places in an a-spatial manner, the internal representations generated for a placename should not be correlated to the internal representations generated for neighbouring placenames. In our experiments, we use spatial autocorrelation to reject this null hypothesis and thus sustain our initial conjecture. 

In this chapter, we establish a novel framework for the study of geospatial mechanistic interpretability, using spatial analysis \cite{osullivan2010geographic} as a tool to explore the geographical aspects of the internal workings of LLMs. In turn, that might provide us with a deeper understanding of geographic bias \cite{decoupes2024evaluation} and diversity \cite{agile-giss-5-38-2024} in LLMs, their ability to reproduce geographical knowledge \cite{mai2023opportunitieschallengesfoundationmodels, roberts2023gpt4geolanguagemodelsees, 10.1145/3589132.3625625} and spatial reasoning \cite{li2024advancing}, and ultimately contribute to AI safety \cite{bereska2024mechanisticinterpretabilityaisafety}. 
The code and data necessary to replicate our experiments are available through our GitHub repository.\footnote{\url{https://github.com/sdesabbata/geospatial-mechanistic-interpretability}}

\section{Probing}
\label{sec:probing}

\subsection{Introduction to probing}

In this chapter, we use the term \emph{probing} to refer to approaches that extract and analyse the activations of an LLM while a prompt is being processed. However, we acknowledge that the term is also sometimes used more colloquially to refer to output-based evaluations or behavioural analyses that focus on the outputs of the models, such as the text produced as an answer to a question in the prompt \cite{hobbhahn2022investigating}.

Probing techniques have emerged as a crucial tool, providing insights into the internal workings of neural networks and, more specifically, LLMs. These techniques aim to unveil the latent knowledge embedded within the internal representations learned by these models. By designing tasks that test specific properties or capabilities of a model, probing enables researchers to assess how well the model encodes and uses various types of linguistic and conceptual knowledge \cite{wallace2019nlp,kim2019probing,belinkov2022probing}.

Probing acts as a diagnostic tool to understand the inner workings of LLMs. Models are presented with controlled tasks that highlight specific functions or internal representations within their architecture. This approach has been instrumental in understanding how LLMs process language, capturing relationships that extend beyond surface-level patterns, where a feature is encapsulated by single neurons, which activate in relation to specific concepts or inputs. It helps researchers investigate not only what LLMs know but also how they process, store, and manipulate information internally. 

A key aspect of probing techniques is their non-intrusive and architecture-agnostic nature. They typically leverage the model in its inference mode, thus without requiring any additional fine-tuning or training, in order to reveal the internal mechanisms that drive decision-making within the model \cite{koto2021discourse,arps2022probing}. 
One common form of LLM probing is through prompting, where carefully crafted inputs are designed to encourage the model to demonstrate its knowledge or capabilities in specific areas \cite{feng-etal-2023-pretraining}. 
By using targeted prompts or inputs, probing techniques can assess a model's knowledge in various areas, such as language understanding, factual knowledge, or even hidden biases \cite{vulic2020probing}. 

Probing techniques often employ targeted tasks that reveal how well the internal representations capture specific linguistic aspects or concepts. A typical probing process involves extracting the model's internal representations while the model is processing a specifically selected set of inputs, as illustrated in Figure~\ref{fig:framework}. A lightweight model (e.g., a linear regression) is then trained to perform a specific task, such as estimating the latitude and longitude of the place mentioned in the input sentence based on the internal representation extracted from the model. If the lightweight model performs well on the task, it suggests that the LLM's internal representations contain the relevant information. For example, if a linear regression can reliably predict the latitude and longitude of a place based on the internal representations extracted from the LLM while it processes a sentence about that place, we can infer that the LLM has encoded some form of geographical awareness in its internal parameters. These types of probes provide insights into how well the LLM has learned to encode fundamental properties, thus offering explainability about both its strengths and limitations in handling basic linguistic properties.

More complex probes can be designed to assess deeper, more abstract properties within an LLM. For example, a probe could be used to determine whether an LLM captures the hierarchical structure of a sentence by training a classifier to distinguish between different syntactic structures, such as identifying clauses or phrases within a sentence \cite{alleman2021syntacticperturbationsrevealrepresentational}. Another example is probing whether an LLM understands semantic relationships between entities \cite{chanin2024identifyinglinearrelationalconcepts}, such as distinguishing between ``Paris is the capital of France'' versus ``France is the capital of Paris.'' In this case, the probe might involve evaluating how well an LLM can represent such relational information by training a classifier to predict entity relationships based on an LLM's internal representations. If the classifier can accurately distinguish between correct and incorrect relationships, it suggests that the LLM has encoded knowledge about semantic roles and relations. By probing for these high-level concepts, researchers can assess an LLM's ability to process and represent more complex aspects of language and knowledge beyond surface-level patterns.

\subsection{Probing for geographical information}
Probing techniques have been used to evaluate the extent to which the internal representations of LLMs, when prompted with geographical entities, can be correlated to information about those geographical entities. For example,
Lietard et al.~\cite{lietard2021language} examined how well the internal representations generated by BERT-like language models \cite{devlin2019bertpretrainingdeepbidirectional} for 3,527 cities and 249 countries and territories can be correlated to their geographical coordinates, population size and neighbouring countries. They found that larger models perform better at encoding geographical information.
Gurnee and Tegmark~\cite{gurneeLanguageModelsRepresent2024} showed how a linear probing technique can be used to predict the geographical coordinates from internal representations within a certain geographical scale and accuracy, concluding that LLMs can learn internal representations of geographical information which are at least partially linear. 
Godey et al.~\cite{godeyScalingLawsGeographical2024} applied the same methodology to a broader set of smaller models, showing how geographical information can be extracted from a wide range of model sizes and that performance is correlated to the amount of geographical information present in the training data. Chen et al.~\cite{chen2023correlationlargelanguagemodels} further expanded on those approaches by introducing multilayer feed-forward neural networks as non-linear probes, aiming to better account for complex internal representations. 

However, the probing techniques outlined above have so far focused on a-spatial statistical modelling, correlating the internal representations to factual information (e.g., population size or approximate location) rather than conducting a spatial analysis of the internal representations. By focusing on the prediction of geographical coordinates from the internal representations, these approaches can provide only a limited insight into the geographical nature of internal representations, and thus limit our ability to explore more complex spatial patterns which might emerge from the internal representations.
As such, in this chapter, we explore the use of spatial analysis \cite{osullivan2010geographic} to explore whether the internal representations of LLMs more broadly follow Tobler's ``first law of geography'' \cite{1aa639e4-d759-3f3c-b072-bba8376952da}. 

\subsection{Spatial analysis of internal representations}
In this section, we explore how spatial autocorrelation \cite{getis2009spatial} can be used to identify interpretable geospatial patterns in internal representations extracted from LLMs. Our methodology involves using placenames alongside their geographical areas as prompts to generate internal representations, which are then correlated with those of nearby placenames using spatial autocorrelation to assess how well the LLM internalises geographical patterns, as illustrated in the bottom-centre component of Figure \ref{fig:framework}.

\subsubsection{Methodology and data}
\label{sec:core-methods-data}

In our experiments, we used three sets of placenames sourced from the GeoNames Free Gazetteer Data repository\footnote{Data collected on July 18th, 2024, from \url{https://download.geonames.org/export/dump/}} representing three areas of broadly similar population sizes: the UK (about 67 million inhabitants), Italy (about 59 million inhabitants), and the four US states commonly associated with the broader New York metropolitan area (New York, New Jersey, Connecticut, and Pennsylvania, totalling about 45.5 million inhabitants). Previous studies \cite{ACHESON2017309} indicate that we can expect a similar high-quality coverage of placenames in those areas. 
The dataset included only populated places (labelled as \texttt{P} in GeoNames) with a recorded population above zero for the UK and the four US states and a population greater than 500 for Italy, resulting in a total of 6,294 placenames for the UK, 9,959 for Italy, and 4,090 for the four US states.

While placename completeness might not be a severe concern in this case, placename ambiguity can impact our experimental setup. Using placenames without further information would likely result in insufficient information for the LLM to correctly interpret the place being referred to due to geo-geo ambiguity (e.g., there is a place called Liverpool in the state of New York and a place called Verona in the state of New Jersey), geo-non-geo ambiguity (e.g., there are 77 places called San Lorenzo in Italy, and the name can also refer to a person from history) and metonymic usage (e.g., using the word ``Westminster'' to refer to the UK government). As such, we decided to input the placenames into the LLM, constructing prompts as follows:

\begin{itemize}
    \item ``\texttt{[placename], [country]}'' for the UK, using the full names of the four countries of the UK (England, Scotland, Wales, and Northern Ireland) based on the value of \texttt{admin1 code} from GeoNames;
    \item ``\texttt{[placename], [province]}'' for Italy, using the full names of the Italian provinces based on the value of \texttt{admin2 code} from GeoNames;
    \item ``\texttt{[placename], [state]}'' for the four US states, using the full names of the states based on the value of \texttt{admin1 code} from GeoNames.
\end{itemize}

We query the LLM using the prompts as described above, capturing the internal representations from different layers of the model for each prompt. Specifically, we rely on \texttt{Mistral-7B-Instruct-v0.2},\footnote{\url{https://huggingface.co/mistralai/Mistral-7B-Instruct-v0.2}} a 7-billion parameter model fine-tuned to follow instruction-based tasks \cite{jiang2023mistral,ouyang2022training}, and we extract the 4,096 post-attention normalised activations of the transformer layers 7, 15, and 31 (counting from zero). These layers were selected to gain an early (token-level patterns), middle (contextual understanding), and late (high-level abstraction) internal representation from the model \cite{templeton2024scaling, gurneeLanguageModelsRepresent2024,elhage2022toymodelssuperposition}.
We then apply \emph{mean pooling} \cite{gholamalinezhad2020pooling}, averaging the token-level activations across each prompt and layer to produce condensed activations.
Thus, each placename is now associated with 4,096 condensed activations per layer, encompassing the internal representations that the model associates with that placename at different processing stages.

To assess whether the LLM's internal representations capture the geographical nature of the places referred to by the input prompts, we need to test whether these condensed activations have similar values for placenames referring to places which are near to one another in geographical space. We thus join the condensed activations generated by each placename with its geographical coordinates and calculate both the global and local Moran's $I$ \cite{af54b142-01be-3f96-af47-1730365d8376} as indicators of spatial autocorrelation \cite{osullivan2010geographic,getis2009spatial}. The global Moran's $I$ indicator is calculated separately for each one of the three areas to assess whether spatial autocorrelation is present in at least one of them, whereas the local clusters are identified by calculating the local Moran's $I$ indicator on the three areas combined, to ensure that a cluster's significance is assessed with regard to all the activations extracted from a layer.

Based on the results presented by previous studies \cite{gurneeLanguageModelsRepresent2024, godeyScalingLawsGeographical2024}, we expect the values to display spatial autocorrelation with values changing across the main cardinal directions. As other authors have been able to use both prompting and linear probes to retrieve the latitude and longitude coordinates from internal representations, it seems probable that LLMs store that information in a manner that leads the values to change linearly as the location of the places changes across the cardinal directions\footnote{It is important to note that LLMs handle numbers in the input query and output answer as textual tokens, which impacts their performance in answering mathematical questions \cite{singh2024tokenizationcountsimpacttokenization}, although recent developments have demonstrated huge improvements in this field through approaches focusing on test-time compute \cite{snell2024scalingllmtesttimecompute}.}. Moreover, due to the inclusion of state or province names in the prompt, we expect to see at least some degree of similarity for values within the same area. However, if the LLM is capable of evoking more nuanced geographical information processing -- or, to be more precise, if the training has produced weights that can link the input placenames to outputs referring to geographical concepts or areas at a different scale(s) -- we should observe spatial autocorrelation with clusters of similar values outlining areas at a different scale compared to the state or province names used in the prompts, or the placenames themselves.

\begin{figure}
    \centering
    \includegraphics[width=1\linewidth]{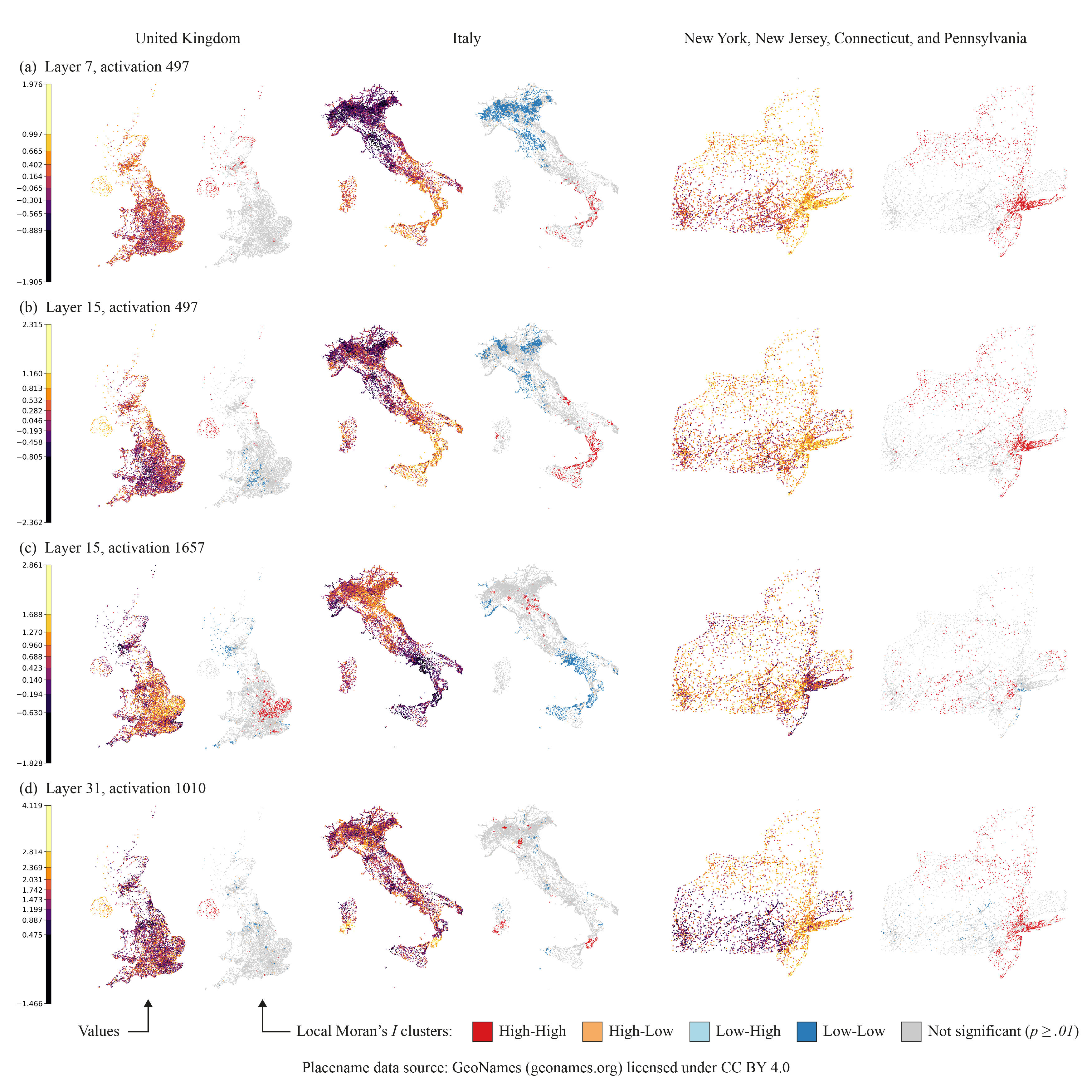}
    \caption{Activations captured for the input placenames at different layers of the LLM (left for each region) and their local spatial autocorrelation (local Moran's $I$ clusters, $p<.01$, right for each region), illustrating the polysemantic nature of its internal representations. Two neurons at layers 7 (a) and 15 (b) show high values for the State of New York and Northern Ireland and very low values for northern Italy. A neuron at layer (c) 15 shows high values for several UK cities. A neuron at layer 31 (d) shows high values for the State of New York and Northern Ireland and diverse values for provinces in Italy.}
    \label{fig:results-map-activations}
\end{figure}

\subsubsection{Results}

Our results illustrate a glimpse of all the aspects mentioned above. While discussing all the 1,841 (14.98\%) neurons across layers 7, 15, and 31 that display a significant autocorrelation ($p<.01$ and Moran's $I \geq 0.3$) is beyond the scope of this chapter, some examples are reported in Figure~\ref{fig:results-map-activations}. 

For instance, Figure~\ref{fig:results-map-activations}(a) illustrates the internal representations of the neuron with the second highest spatial autocorrelation at layer 7. Very high values highlight areas whose name was used as part of the prompts (e.g., ``\texttt{New York}'' as \texttt{[state]} in the input for placenames in the State of New York or ``\texttt{Northern Ireland}'' as \texttt{[country]} for placenames in Northern Ireland, as outlined in Section~\ref{sec:core-methods-data}). At the same time, a clear pattern of increasing values can be seen by moving from the north of Italy to the south, despite prompts not including explicit references to southern, central or northern Italy, only the name of the place and province. This indicates that the LLM might be encoding broader regions or geographical locations as part of the internal representation of place and province names, which could be linked to LLMs' ability to generate latitude and longitude values for input placenames \cite{mai2023opportunitieschallengesfoundationmodels, 10.1145/3589132.3625625}. A very similar pattern is also displayed by the neuron with the sixth-highest spatial autocorrelation at layer 15, as shown in Figure~\ref{fig:results-map-activations}(b), which indicates some form of consistency across layers. However, the neuron with the second highest spatial autocorrelation at layer 15 might be introducing some level of complexity that goes beyond the linearity of cartographic coordinates or named areas. As shown in Figure~\ref{fig:results-map-activations}(c), that particular neuron seems to reveal particularly high values in East Anglia and the East Midlands, with lower values in larger cities such as London, Birmingham, Liverpool and Manchester in the UK, as well as New York City and Philadelphia in the US. At the same time, the neuron also seems to display a gradual change from the north of Italy to the south, as in Figure~\ref{fig:results-map-activations}(a) and (b), without a clear pattern related to major cities. Figure~\ref{fig:results-map-activations}(b) and (c) illustrate how different neurons represent different information within the same layer. Interestingly, the final layer seems to focus back to local or broad-stroke geographies, as illustrated in Figure~\ref{fig:results-map-activations}(d) by the neuron with the highest spatial autocorrelation at layer 31, which shows low values for Pennsylvania and high values for the State of New York, generally low values for Great Britain and high in Northern Ireland, and a patchwork of high and low values across Italy.

Importantly, the neurons showing high spatial autocorrelation do not seem to behave in a monosemantic (one-meaning) manner, where they would output high values for one area or concept only. Rather, they can behave in a polysemantic (multiple-meaning) manner, where the same neuron (same row in Figure~\ref{fig:results-map-activations}) outputs high values for different areas or concepts. That seems to indicate that the superposition hypothesis (introduced below) might hold for geographical information. Therefore, in the next section, we explore a mechanistic interpretability approach aimed at addressing this limitation by decomposing these polysemantic structures into more interpretable, monosemantic features. In particular, we explore the use of sparse autoencoders to identify specific directions in the LLM's internal representation space that correspond to distinct geographical entities and concepts, allowing us to better understand how LLMs process geographical information.

\section{Mechanistic interpretability}

\subsection{Introduction to mechanistic interpretability}
Mechanistic interpretability seeks to understand how LLMs process and encode information by breaking down their internal representations into more interpretable features. 
A key concept in mechanistic interpretability is the \emph{superposition hypothesis} \cite{templeton2024scaling,elhage2022toymodelssuperposition, bricken2023monosemanticity}, which suggests that neural networks often need to represent more features than they have neurons. This means multiple features can be encoded in overlapping or shared neurons, making it difficult to clearly interpret individual neurons. To address this, researchers employ methods like sparse autoencoders, which attempt to disentangle the model's internal representations into a set of more interpretable, sparse features \cite{bricken2023monosemanticity}. The concept of \emph{monosemanticity} instead refers to the idea that certain neurons in the network consistently correspond to a single interpretable concept. 

In practice, many LLM neurons are polysemantic, meaning they can represent multiple, sometimes unrelated, concepts depending on the context. Scaling studies have shown that larger models tend to exhibit more monosemantic features, which improves interpretability as each feature is more likely to correspond to a unique concept \cite{templeton2024scaling}.

Sparse autoencoders \cite{ng2011sparse} play a crucial role in this process by decomposing the internal representations of models into sparse, interpretable directions of a multi-dimensional conceptual space. These directions correspond to distinct features, allowing researchers to better understand the relationship between models' inputs and outputs. In recent work, the use of sparse autoencoders has shown that even small transformer models, such as those with just a single MLP layer, can be decomposed into thousands of features, each capturing a different, often interpretable, property of the input data \cite{bricken2023monosemanticity}.

Overall, mechanistic interpretability provides a framework for understanding the inner workings of LLMs. By exploring notions and techniques like superposition, monosemanticity, and sparse autoencoders, researchers can gain deeper insights into how these models encode and manipulate complex features, ultimately improving our ability to interpret and control their behaviour.

\subsection{Mechanistic interpretability via sparse autoencoders and spatial analysis}

In this section, we describe our approach to extracting geospatially interpretable features using a sparse autoencoder \cite{ng2011sparse} and spatial autocorrelation \cite{osullivan2010geographic,getis2009spatial}. That is illustrated in Figure \ref{fig:framework}, where the features obtained from the sparse autoencoder in the bottom-right component are used as input for a spatial autocorrelation analysis in the bottom-centre component. The primary goal is to understand how LLMs encode geographical information and whether these internal representations display geospatial patterns.

\subsubsection{Methodology and data}
For the preliminary experiments presented in this chapter, we begin by training a sparse autoencoder on the condensed activations described in Section \ref{sec:core-methods-data}. The sparse autoencoder is designed to decompose the condensed activations into a set of sparse, interpretable features \cite{templeton2024scaling,bricken2023monosemanticity}, allowing us to identify the directions in the LLM's internal representation space that correspond to specific geographical entities or concepts.

We followed an approach similar to the one proposed by Templeton et al.~\cite{templeton2024scaling}. We focused on the middle layer (i.e., layer 15) and used its condensed activations to train a sparse autoencoder with a single encoding layer that expanded 4,096 input condensed activations to 32,768 embeddings (i.e., eight times the input size) and a single decoding layer to reduce them back to 4,096 reconstructed values. However, instead of relying on including an L1 penalty to encourage sparsity, we employed the TopK with ReLU activation function proposed by Gao et al.~\cite{gao2024scalingevaluatingsparseautoencoders}, which retains only the $k$ largest embeddings, setting the rest to zero. We trained four models with $k$ values of 1,024, 2,048, 4,096 and 8,192 (to test different levels of sparsity) over 300 epochs using the whole dataset (as we did not seek to build a generalisable model) in mini-batches of 32 cases in randomly shuffled order. Based on the final training loss, we selected the model with $k$ equal to 2,048 as the best model. Finally, we used the encoder from this model to encode each 4,096 condensed activations into 32,768 features.

\begin{figure}
    \centering
    \includegraphics[width=1\linewidth]{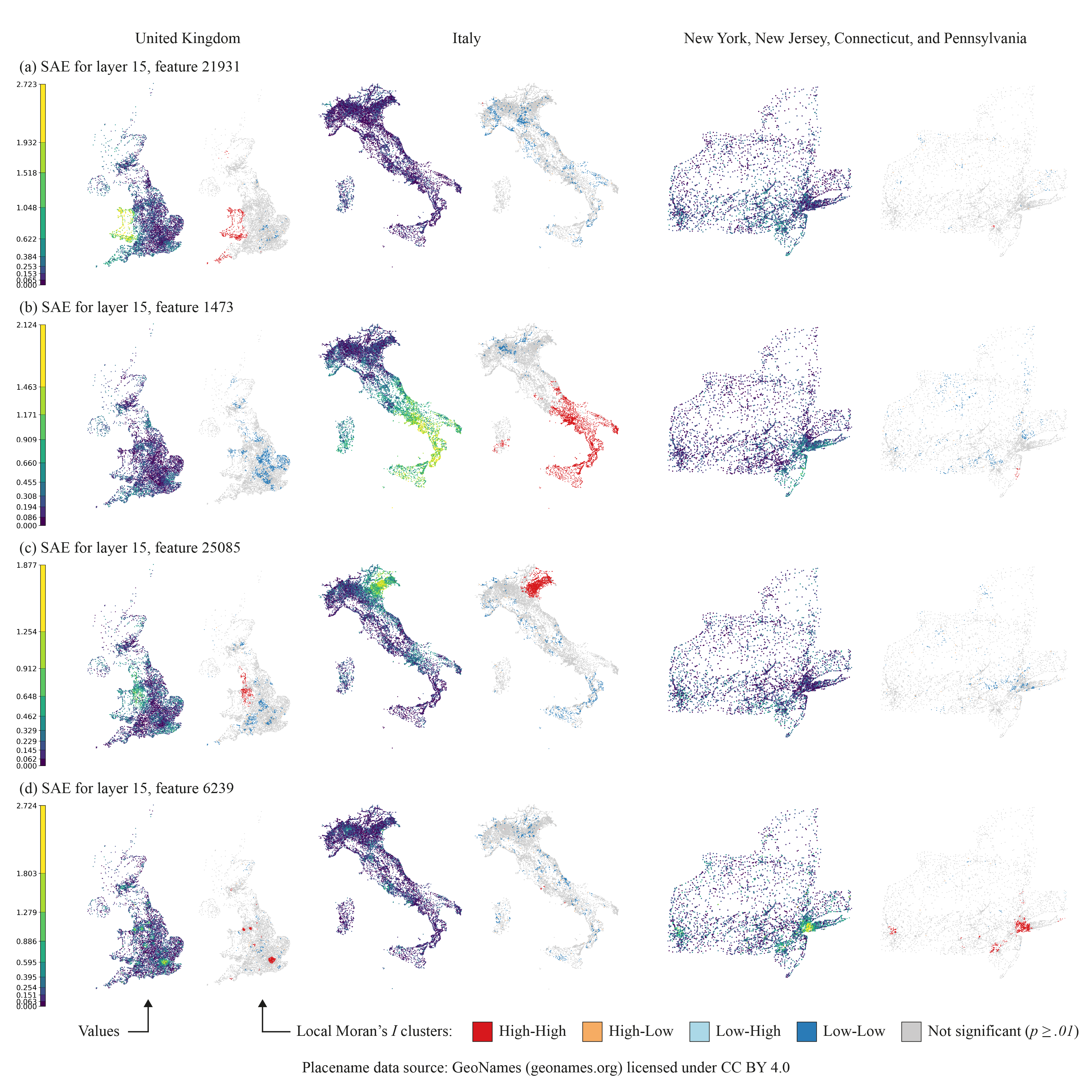}
    \caption{Features extracted from layer 15 through a sparse autoencoder (left for each region) and their local spatial autocorrelation (local Moran's $I$ clusters, $p<.01$, right for each region): (a) Wales as a region part of prompt; (b) south of Italy as a region activating a seemingly monosemantic feature; (c) north-east of Italy and north-west of England as regions activating a seemingly polysemantic feature; and (d) a representation of ``city'' highlighting New York City and London, amongst others.}
    \label{fig:results-map-sae-l15}
\end{figure}

Once the features were extracted, we proceeded as in the previous experiment, linking back the features to the geographical coordinates and calculating the global and local Moran's $I$ \cite{af54b142-01be-3f96-af47-1730365d8376} indicators of spatial autocorrelation \cite{osullivan2010geographic,getis2009spatial}.

\subsubsection{Results}

{Our preliminary results illustrate how the features extracted by the sparse autoencoder seem to display a clearer monosemantic and spatially coherent nature.}

For instance, Figure~\ref{fig:results-map-sae-l15}(a) illustrates an example of a feature encoding a region that was part of the prompt, in this case, Wales. Figure~\ref{fig:results-map-sae-l15}(b) and (c) are, instead, examples of features displaying strong spatial autocorrelation and high values across areas that were not explicitly part of input prompts. While Italian provinces were used in the prompts for Italian placenames, the feature in Figure~\ref{fig:results-map-sae-l15}(b) encodes the whole of southern Italy, and the feature in Figure~\ref{fig:results-map-sae-l15}(c) encodes the north-east areas of the country. At the same time, while the feature illustrated in \ref{fig:results-map-sae-l15}(b) seems to be strongly monosemantic -- with very high values in the south of Italy and low values in the centre and north of Italy, as well as across the UK and the US states included in the analysis -- the feature illustrated in \ref{fig:results-map-sae-l15}(c) displays high values for both the north-east of Italy (especially around Venice) and the north-west of England (especially around Liverpool). A connection might be drawn around the historical importance of Liverpool and Venice as ports and areas of industrial development. However, that would exclude other places with comparable port and industrial histories that are part of the maps but do not display high values, such as Genova. A clearer display of an important geographical concept (i.e., ``urban'') is instead illustrated in \ref{fig:results-map-sae-l15}(d), where high values are visible for several large cities, including New York City, Philadelphia, Pittsburg, London, Manchester and Milano. 

It is noteworthy that only 67 of the 32,768 features (0.2\%) displayed a significant spatial autocorrelation ($p < .01$ and Moran's $I \geq 0.3$). This indicates that LLMs may encode some degree of geographical information, but this information is represented in a sparse and diffuse manner across the model's internal layers. The small number of spatially coherent features could be a result of the superposition hypothesis, where geographical concepts are entangled with other unrelated concepts, making them more difficult to isolate. However, the presence of features with significant spatial autocorrelation also suggests that LLMs are capable of capturing and representing geospatial patterns, although these patterns might only manifest in specific layers or under particular conditions. Further investigation into these features could reveal more about how LLMs generalise geographical knowledge and how they differentiate between regions based on learned associations. Moreover, 99.53\% of features generated by the encoder component of the sparse autoencoder trained for our preliminary experiments always generate a zero value for all input placenames, leaving 0.27\% of features displaying minimal or not significant spatial autocorrelation. That suggests that further work is necessary to improve the quality of the decomposition, as different choices in the extracted activations, the construction of the sparse autoencoder (e.g., size of the ``bottleneck'' or approach to sparsity) or training might provide better insights or that expanding the internal representations to such a large number of features might not be necessary when focusing on a specific subject or small datasets compared to broader studies \cite{templeton2024scaling}.

\section{Discussion and Conclusions}

This chapter proposes a novel framework as a path towards the study of geospatial mechanistic interpretability of LLMs and, thus, an entry point into an array of new research question \cite{sharkey2025openproblemsmechanisticinterpretability}. Our preliminary experiments provide insights into the internal representations that LLMs use to handle geographical information, leading us to two key findings. 

First, the internal representations handling geographical information in LLMs seem to be often distributed across many polysemantic neurons rather than being monosemantically represented by a single neuron. That is, a geographical entity or concept may be encoded in a non-trivial manner, activating multiple neurons at the same time, and a single neuron can encode multiple, unrelated geographical entities or concepts -- possibly alongside other information -- which is consistent with the superposition hypothesis \cite{elhage2022toymodelssuperposition}. The nature of these internal representations makes it difficult to isolate and interpret individual geographical entities and concepts. This finding raised important questions about the geospatial mechanistic interpretability of LLMs and their ability to generalise geographical knowledge, which led us to explore the use of sparse autoencoders \cite{ng2011sparse} and our second finding.

Second, the combined use of sparse autoencoders and spatial analysis seems capable of rendering the internal representations handling geographical information in LLMs interpretable. The use of sparse autoencoders allowed us to disentangle some of the polysemantic structures into more interpretable, monosemantic features. By applying spatial autocorrelation to these disentangled features, we observe clearer geospatial patterns, which highlight the potential of sparse autoencoders to improve the interpretability of LLMs for geospatial tasks. 

At the same time, several geographical questions remain open, which will require further investigation: How does placename ambiguity impact LLMs' learning and interpretability? What scale(s) of geographical relationships and effects are encoded, and at which depth(s) of the models? How do geographical relationships compare to and interact with other types of relationships (e.g., social, economic, or historical)? 

Our future work will explore those questions for LLMs and foundation models \cite{mai2023opportunitieschallengesfoundationmodels,bommasani2022opportunitiesrisksfoundationmodels} more broadly. We aim to further develop our geospatial mechanistic interpretability framework, for instance, by exploring further methods in spatial analysis \cite{osullivan2010geographic}, the discovery of feature circuits \cite{wang2022interpretabilitywildcircuitindirect, marks2024sparsefeaturecircuitsdiscovering} and different prompting and sparse autoencoder training strategies to better understand the internal representations activated in different contexts. Moreover, future work could explore how different training strategies, model architectures, and datasets influence the encoding of geographical information, as integrating external geographical knowledge bases or geographical embeddings could also improve the LLMs' ability to handle geographical entities, relationships and concepts. The study of the internal representations generated for different languages, alternative and vernacular placenames \cite{Jones01102008}, and place nouns could also provide further insights into the internal geographies of LLMs.

Our approach holds the potential to substantially expand our understanding of how LLMs and foundation models handle geographical information. This is an essential step in developing robust and reliable tools based on foundation models for geography and possibly even developing the ``artificial GIS analyst that passes a domain-specific Turing Test by 2030'' imagined by Janowicz et al.~\cite{doi:10.1080/13658816.2019.1684500}, as Chen et al.~\cite{chen2023correlationlargelanguagemodels} illustrated how the quality of the internal representations of geographical information can influenced an LLM's performance on geospatial tasks. Facilitated by the novel framework presented in this chapter, the study of geospatial mechanistic interpretability might lead to a better understanding of how LLMs internally relate geographical information to cultural, socio-economic, political, or environmental information, thus complementing output-based evaluations focusing on LLM bias \cite{decoupes2024evaluation} and diversity \cite{agile-giss-5-38-2024}, as well as quality in reproducing geographical knowledge \cite{mai2023opportunitieschallengesfoundationmodels, roberts2023gpt4geolanguagemodelsees, 10.1145/3589132.3625625} and spatial reasoning \cite{li2024advancing}, and thus contributing to AI safety \cite{bereska2024mechanisticinterpretabilityaisafety}.

\section*{Acknowledgments}
The authors would like to thank Univ.-Prof. Dr. Krzysztof Janowicz, Dr. Rui Zhu and the anonymous reviewers for their valuable comments, which helped us shape this chapter. This research used the ALICE High Performance Computing Facility at the University of Leicester.

\bibliographystyle{Vancouver}
\bibliography{main}

\begin{thebibliography}{10}

\bibitem{mai2023opportunitieschallengesfoundationmodels}
Mai G, Huang W, Sun J, Song S, Mishra D, Liu N, et~al.
\newblock On the Opportunities and Challenges of Foundation Models for GeoAI
  (Vision Paper).
\newblock ACM Trans Spatial Algorithms Syst. 2024 Jul;10(2).

\bibitem{doi:10.1080/17538947.2024.2353122}
Wang S, Hu T, Xiao H, Li Y, Zhang C, Ning H, et~al.
\newblock GPT, large language models (LLMs) and generative artificial
  intelligence (GAI) models in geospatial science: a systematic review.
\newblock International Journal of Digital Earth. 2024;17(1):2353122.

\bibitem{Hochmair_2024}
Hochmair HH, Juhász L, Kemp T.
\newblock Correctness Comparison of ChatGPT‐4, Gemini, Claude‐3, and
  Copilot for Spatial Tasks.
\newblock Transactions in GIS. 2024 Aug.

\bibitem{xu2024evaluatinglargelanguagemodels}
Xu L, Zhao S, Lin Q, Chen L, Luo Q, Wu S, et~al.. Evaluating Large Language
  Models on Spatial Tasks: A Multi-Task Benchmarking Study; 2024.
\newblock Available from: \url{https://arxiv.org/abs/2408.14438}.

\bibitem{cohn2023dialecticallanguagemodelevaluation}
Cohn AG, Hernandez-Orallo J. Dialectical language model evaluation: An initial
  appraisal of the commonsense spatial reasoning abilities of LLMs; 2023.
\newblock Available from: \url{https://arxiv.org/abs/2304.11164}.

\bibitem{https://doi.org/10.4230/lipics.cosit.2024.28}
Cohn AG, Blackwell RE.
\newblock {Evaluating the Ability of Large Language Models to Reason About
  Cardinal Directions}.
\newblock In: Adams B, Griffin AL, Scheider S, McKenzie G, editors. 16th
  International Conference on Spatial Information Theory (COSIT 2024). vol. 315
  of Leibniz International Proceedings in Informatics (LIPIcs). Dagstuhl,
  Germany: Schloss Dagstuhl -- Leibniz-Zentrum f{\"u}r Informatik; 2024. p.
  28:1-28:9.

\bibitem{li2024advancing}
Li F, Hogg DC, Cohn AG.
\newblock Advancing spatial reasoning in large language models: an in-depth
  evaluation and enhancement using the StepGame benchmark.
\newblock In: Proceedings of the Thirty-Eighth AAAI Conference on Artificial
  Intelligence and Thirty-Sixth Conference on Innovative Applications of
  Artificial Intelligence and Fourteenth Symposium on Educational Advances in
  Artificial Intelligence. AAAI'24/IAAI'24/EAAI'24. AAAI Press; 2024. .

\bibitem{Hu24092024}
Hu X, Kersten J, Klan F, Farzana SM.
\newblock Toponym resolution leveraging lightweight and open-source large
  language models and geo-knowledge.
\newblock International Journal of Geographical Information Science.
  2024;0(0):1-28.

\bibitem{li2023autonomous}
Li Z, Ning H.
\newblock Autonomous GIS: the next-generation AI-powered GIS.
\newblock International Journal of Digital Earth. 2023;16(2):4668-86.

\bibitem{zhang2023geogptunderstandingprocessinggeospatial}
Zhang Y, Wei C, He Z, Yu W.
\newblock GeoGPT: An assistant for understanding and processing geospatial
  tasks.
\newblock International Journal of Applied Earth Observation and
  Geoinformation. 2024;131:103976.

\bibitem{zhang2024bb}
Zhang Y, Wang Z, He Z, Li J, Mai G, Lin J, et~al.
\newblock BB-GeoGPT: A framework for learning a large language model for
  geographic information science.
\newblock Information Processing \& Management. 2024;61(5):103808.

\bibitem{doi:10.1080/15230406.2024.2404868}
Zhang Y, He Z, Li J, Lin J, Guan Q, Yu W.
\newblock MapGPT: an autonomous framework for mapping by integrating large
  language model and cartographic tools.
\newblock Cartography and Geographic Information Science. 2024;0(0):1-27.

\bibitem{singh2024geollm}
Singh S, Fore M, Stamoulis D.
\newblock GeoLLM-Engine: A Realistic Environment for Building Geospatial
  Copilots.
\newblock In: Proceedings of the IEEE/CVF Conference on Computer Vision and
  Pattern Recognition; 2024. p. 585-94.

\bibitem{tan2023promiseschallengesmultimodalfoundation}
Tan C, Cao Q, Li Y, Zhang J, Yang X, Zhao H, et~al.. On the Promises and
  Challenges of Multimodal Foundation Models for Geographical, Environmental,
  Agricultural, and Urban Planning Applications; 2023.
\newblock Available from: \url{https://arxiv.org/abs/2312.17016}.

\bibitem{zhu2024plangptenhancingurbanplanning}
Zhu H, Zhang W, Huang N, Li B, Niu L, Fan Z, et~al.. PlanGPT: Enhancing Urban
  Planning with Tailored Language Model and Efficient Retrieval; 2024.
\newblock Available from: \url{https://arxiv.org/abs/2402.19273}.

\bibitem{doi:10.1080/13658816.2023.2266495}
Hu Y, Mai G, Cundy C, Choi K, Lao N, Liu W, et~al.
\newblock Geo-knowledge-guided GPT models improve the extraction of location
  descriptions from disaster-related social media messages.
\newblock International Journal of Geographical Information Science.
  2023;37(11):2289-318.

\bibitem{Roberts_2024_CVPR}
Roberts J, L\"uddecke T, Sheikh R, Han K, Albanie S.
\newblock Charting New Territories: Exploring the Geographic and Geospatial
  Capabilities of Multimodal LLMs.
\newblock In: Proceedings of the IEEE/CVF Conference on Computer Vision and
  Pattern Recognition (CVPR) Workshops; 2024. p. 554-63.

\bibitem{feng2024citygptempoweringurbanspatial}
Feng J, Du Y, Liu T, Guo S, Lin Y, Li Y. CityGPT: Empowering Urban Spatial
  Cognition of Large Language Models; 2024.
\newblock Available from: \url{https://arxiv.org/abs/2406.13948}.

\bibitem{fulman2024distortionsjudgedspatialrelations}
Fulman N, Memduhoğlu A, Zipf A.
\newblock Distortions in Judged Spatial Relations in Large Language Models.
\newblock The Professional Geographer. 2024;76(6):703-11.

\bibitem{feng2024move}
Feng S, Lyu H, Li F, Sun Z, Chen C.
\newblock Where to move next: Zero-shot generalization of llms for next poi
  recommendation.
\newblock In: 2024 IEEE Conference on Artificial Intelligence (CAI). IEEE;
  2024. p. 1530-5.

\bibitem{ilyankou2024sentencetransformerslearnquasigeospatial}
Ilyankou I, Lipani A, Cavazzi S, Gao X, Haworth J. Do Sentence Transformers
  Learn Quasi-Geospatial Concepts from General Text?; 2024.
\newblock Available from: \url{https://arxiv.org/abs/2404.04169}.

\bibitem{roberts2023gpt4geolanguagemodelsees}
Roberts J, L{\"u}ddecke T, Das S, Han K, Albanie S. GPT4GEO: How a Language
  Model Sees the World's Geography; 2023.
\newblock Available from: \url{https://arxiv.org/abs/2306.00020}.

\bibitem{10.1145/3589132.3625625}
Bhandari P, Anastasopoulos A, Pfoser D.
\newblock Are Large Language Models Geospatially Knowledgeable?
\newblock In: Proceedings of the 31st ACM International Conference on Advances
  in Geographic Information Systems. SIGSPATIAL '23. New York, NY, USA:
  Association for Computing Machinery; 2023. .

\bibitem{decoupes2024evaluation}
Decoupes R, Interdonato R, Roche M, Teisseire M, Valentin S.
\newblock Evaluation of Geographical Distortions in Language Models: A Crucial
  Step Towards Equitable Representations.
\newblock arXiv preprint arXiv:240417401. 2024.

\bibitem{agile-giss-5-38-2024}
Liu Z, Janowicz K, Currier K, Shi M.
\newblock Measuring Geographic Diversity of Foundation Models with a Natural
  Language–based Geo-guessing Experiment on GPT-4.
\newblock AGILE: GIScience Series. 2024;5:38.

\bibitem{liuMakingGeographicSpace}
Liu Z, Currier K, Janowicz K.
\newblock Making {{Geographic Space Explicit In Probing Multimodal Large
  Language Models For Cul-Tural Subjects}}.
\newblock In: Global {{AI Cultures}} Workshop of {{ICLR}} 2024; 2024. .

\bibitem{BERRAGAN2024102121}
Berragan C, Singleton A, Calafiore A, Morley J.
\newblock Mapping Great Britain's semantic footprints through a large language
  model analysis of Reddit comments.
\newblock Computers, Environment and Urban Systems. 2024;110:102121.

\bibitem{lecun2015deep}
LeCun Y, Bengio Y, Hinton G.
\newblock Deep learning.
\newblock nature. 2015;521(7553):436-44.

\bibitem{nair2010rectified}
Nair V, Hinton GE.
\newblock Rectified linear units improve restricted boltzmann machines.
\newblock In: Proceedings of the 27th international conference on machine
  learning (ICML-10); 2010. p. 807-14.

\bibitem{vaswani2023attentionneed}
Vaswani A, Shazeer N, Parmar N, Uszkoreit J, Jones L, Gomez AN, et~al.
\newblock Attention is All you Need.
\newblock In: Guyon I, Luxburg UV, Bengio S, Wallach H, Fergus R, Vishwanathan
  S, et~al., editors. Advances in Neural Information Processing Systems.
  vol.~30. Curran Associates, Inc.; 2017. .

\bibitem{6472238}
Bengio Y, Courville A, Vincent P.
\newblock Representation Learning: A Review and New Perspectives.
\newblock IEEE Transactions on Pattern Analysis and Machine Intelligence.
  2013;35(8):1798-828.

\bibitem{templeton2024scaling}
Templeton A, Conerly T, Marcus J, Lindsey J, Bricken T, Chen B, et~al.
\newblock Scaling Monosemanticity: Extracting Interpretable Features from
  Claude 3 Sonnet.
\newblock Transformer Circuits Thread. 2024.

\bibitem{lillicrap2020backpropagation}
Lillicrap TP, Santoro A, Marris L, Akerman CJ, Hinton G.
\newblock Backpropagation and the brain.
\newblock Nature Reviews Neuroscience. 2020;21(6):335-46.

\bibitem{ziegler2020finetuninglanguagemodelshuman}
Ziegler DM, Stiennon N, Wu J, Brown TB, Radford A, Amodei D, et~al..
  Fine-Tuning Language Models from Human Preferences; 2020.
\newblock Available from: \url{https://arxiv.org/abs/1909.08593}.

\bibitem{1aa639e4-d759-3f3c-b072-bba8376952da}
Tobler WR.
\newblock A Computer Movie Simulating Urban Growth in the Detroit Region.
\newblock Economic Geography. 1970;46:234-40.

\bibitem{lietard2021language}
Li{\'e}tard B, Abdou M, S{\o}gaard A.
\newblock Do Language Models Know the Way to Rome?
\newblock arXiv preprint arXiv:210907971. 2021.

\bibitem{gurneeLanguageModelsRepresent2024}
Gurnee W, Tegmark M. Language {{Models Represent Space}} and {{Time}}. arXiv;
  2024.

\bibitem{godeyScalingLawsGeographical2024}
Godey N, {de la Clergerie} {\'E}, Sagot B. On the {{Scaling Laws}} of
  {{Geographical Representation}} in {{Language Models}}. arXiv; 2024.

\bibitem{chen2023correlationlargelanguagemodels}
Chen Y, Gan Y, Li S, Yao L, Zhao X. More than Correlation: Do Large Language
  Models Learn Causal Representations of Space?; 2023.
\newblock Available from: \url{https://arxiv.org/abs/2312.16257}.

\bibitem{osullivan2010geographic}
{O’Sullivan} D, Unwin D.
\newblock Geographic information analysis.
\newblock John Wiley \& Sons; 2010.

\bibitem{bereska2024mechanisticinterpretabilityaisafety}
Bereska L, Gavves E. Mechanistic Interpretability for AI Safety -- A Review;
  2024.
\newblock Available from: \url{https://arxiv.org/abs/2404.14082}.

\bibitem{hobbhahn2022investigating}
Hobbhahn M, Lieberum T, Seiler D.
\newblock Investigating causal understanding in LLMs.
\newblock In: NeurIPS ML Safety Workshop; 2022. .

\bibitem{wallace2019nlp}
Wallace E, Wang Y, Li S, Singh S, Gardner M.
\newblock Do NLP models know numbers? probing numeracy in embeddings.
\newblock arXiv preprint arXiv:190907940. 2019.

\bibitem{kim2019probing}
Kim N, Patel R, Poliak A, Wang A, Xia P, McCoy RT, et~al.
\newblock Probing what different NLP tasks teach machines about function word
  comprehension.
\newblock arXiv preprint arXiv:190411544. 2019.

\bibitem{belinkov2022probing}
Belinkov Y.
\newblock Probing classifiers: Promises, shortcomings, and advances.
\newblock Computational Linguistics. 2022;48(1):207-19.

\bibitem{koto2021discourse}
Koto F, Lau JH, Baldwin T.
\newblock Discourse probing of pretrained language models.
\newblock arXiv preprint arXiv:210405882. 2021.

\bibitem{arps2022probing}
Arps D, Samih Y, Kallmeyer L, Sajjad H.
\newblock Probing for constituency structure in neural language models.
\newblock arXiv preprint arXiv:220406201. 2022.

\bibitem{feng-etal-2023-pretraining}
Feng S, Park CY, Liu Y, Tsvetkov Y.
\newblock From Pretraining Data to Language Models to Downstream Tasks:
  Tracking the Trails of Political Biases Leading to Unfair {NLP} Models.
\newblock In: Rogers A, Boyd-Graber J, Okazaki N, editors. Proceedings of the
  61st Annual Meeting of the Association for Computational Linguistics (Volume
  1: Long Papers). Toronto, Canada: Association for Computational Linguistics;
  2023. p. 11737-62.

\bibitem{vulic2020probing}
Vuli{\'c} I, Ponti EM, Litschko R, Glava{\v{s}} G, Korhonen A.
\newblock Probing pretrained language models for lexical semantics.
\newblock In: Proceedings of the 2020 Conference on Empirical Methods in
  Natural Language Processing (EMNLP); 2020. p. 7222-40.

\bibitem{alleman2021syntacticperturbationsrevealrepresentational}
Alleman M, Mamou J, Del~Rio MA, Tang H, Kim Y, Chung S. Syntactic Perturbations
  Reveal Representational Correlates of Hierarchical Phrase Structure in
  Pretrained Language Models; 2021.
\newblock Available from: \url{https://arxiv.org/abs/2104.07578}.

\bibitem{chanin2024identifyinglinearrelationalconcepts}
Chanin D, Hunter A, Camburu OM. Identifying Linear Relational Concepts in Large
  Language Models; 2024.
\newblock Available from: \url{https://arxiv.org/abs/2311.08968}.

\bibitem{devlin2019bertpretrainingdeepbidirectional}
Devlin J, Chang MW, Lee K, Toutanova K.
\newblock {BERT}: Pre-training of Deep Bidirectional Transformers for Language
  Understanding.
\newblock In: Burstein J, Doran C, Solorio T, editors. Proceedings of the 2019
  Conference of the North {A}merican Chapter of the Association for
  Computational Linguistics: Human Language Technologies, Volume 1 (Long and
  Short Papers). Minneapolis, Minnesota: Association for Computational
  Linguistics; 2019. p. 4171-86.
\newblock Available from: \url{https://aclanthology.org/N19-1423/}.

\bibitem{getis2009spatial}
Getis A.
\newblock Spatial autocorrelation.
\newblock In: Handbook of applied spatial analysis: Software tools, methods and
  applications. Springer; 2009. p. 255-78.

\bibitem{ACHESON2017309}
Acheson E, {De Sabbata} S, Purves RS.
\newblock A quantitative analysis of global gazetteers: Patterns of coverage
  for common feature types.
\newblock Computers, Environment and Urban Systems. 2017;64:309-20.

\bibitem{jiang2023mistral}
Jiang AQ, Sablayrolles A, Mensch A, Bamford C, Chaplot DS, Casas Ddl, et~al.
\newblock Mistral 7B.
\newblock arXiv preprint arXiv:231006825. 2023.

\bibitem{ouyang2022training}
Ouyang L, Wu J, Jiang X, Almeida D, Wainwright C, Mishkin P, et~al.
\newblock Training language models to follow instructions with human feedback.
\newblock Advances in neural information processing systems. 2022;35:27730-44.

\bibitem{elhage2022toymodelssuperposition}
Elhage N, Hume T, Olsson C, Schiefer N, Henighan T, Kravec S, et~al.. Toy
  Models of Superposition; 2022.
\newblock Available from: \url{https://arxiv.org/abs/2209.10652}.

\bibitem{gholamalinezhad2020pooling}
Gholamalinezhad H, Khosravi H.
\newblock Pooling methods in deep neural networks, a review.
\newblock arXiv preprint arXiv:200907485. 2020.

\bibitem{af54b142-01be-3f96-af47-1730365d8376}
Moran PAP.
\newblock Notes on Continuous Stochastic Phenomena.
\newblock Biometrika. 1950;37(1/2):17-23.

\bibitem{singh2024tokenizationcountsimpacttokenization}
Singh AK, Strouse D. Tokenization counts: the impact of tokenization on
  arithmetic in frontier LLMs; 2024.
\newblock Available from: \url{https://arxiv.org/abs/2402.14903}.

\bibitem{snell2024scalingllmtesttimecompute}
Snell C, Lee J, Xu K, Kumar A. Scaling LLM Test-Time Compute Optimally can be
  More Effective than Scaling Model Parameters; 2024.
\newblock Available from: \url{https://arxiv.org/abs/2408.03314}.

\bibitem{bricken2023monosemanticity}
Bricken T, Templeton A, Batson J, Chen B, Jermyn A, Conerly T, et~al.
\newblock Towards Monosemanticity: Decomposing Language Models With Dictionary
  Learning.
\newblock Transformer Circuits Thread. 2023.
\newblock
  Https://transformer-circuits.pub/2023/monosemantic-features/index.html.

\bibitem{ng2011sparse}
Ng A, et~al.
\newblock Sparse autoencoder.
\newblock CS294A Lecture notes. 2011;72(2011):1-19.

\bibitem{gao2024scalingevaluatingsparseautoencoders}
Gao L, la~Tour TD, Tillman H, Goh G, Troll R, Radford A, et~al.. Scaling and
  evaluating sparse autoencoders; 2024.
\newblock Available from: \url{https://arxiv.org/abs/2406.04093}.

\bibitem{sharkey2025openproblemsmechanisticinterpretability}
Sharkey L, Chughtai B, Batson J, Lindsey J, Wu J, Bushnaq L, et~al.. Open
  Problems in Mechanistic Interpretability; 2025.
\newblock Available from: \url{https://arxiv.org/abs/2501.16496}.

\bibitem{bommasani2022opportunitiesrisksfoundationmodels}
Bommasani R, Hudson DA, Adeli E, Altman R, Arora S, von Arx S, et~al.. On the
  Opportunities and Risks of Foundation Models; 2022.
\newblock Available from: \url{https://arxiv.org/abs/2108.07258}.

\bibitem{wang2022interpretabilitywildcircuitindirect}
Wang K, Variengien A, Conmy A, Shlegeris B, Steinhardt J. Interpretability in
  the Wild: a Circuit for Indirect Object Identification in GPT-2 small; 2022.
\newblock Available from: \url{https://arxiv.org/abs/2211.00593}.

\bibitem{marks2024sparsefeaturecircuitsdiscovering}
Marks S, Rager C, Michaud EJ, Belinkov Y, Bau D, Mueller A. Sparse Feature
  Circuits: Discovering and Editing Interpretable Causal Graphs in Language
  Models; 2024.
\newblock Available from: \url{https://arxiv.org/abs/2403.19647}.

\bibitem{Jones01102008}
Jones CB, Purves RS, Clough PD, Joho H.
\newblock Modelling vague places with knowledge from the Web.
\newblock International Journal of Geographical Information Science.
  2008;22(10):1045-65.

\bibitem{doi:10.1080/13658816.2019.1684500}
Janowicz K, Gao S, McKenzie G, Hu Y, Bhaduri B.
\newblock GeoAI: spatially explicit artificial intelligence techniques for
  geographic knowledge discovery and beyond.
\newblock International Journal of Geographical Information Science.
  2020;34(4):625-36.

\end{thebibliography}
\end{document}